# A NEW APPROACH OF GRAY IMAGES BINARIZATION WITH THRESHOLD METHODS

Andrei Hossu              Daniela Andone

POLITEHNICA UNIVERSITY OF BUCHAREST
Faculty of Control and Computer Science, ROMANIA,
E-mail: hossu@aii.pub.ro,

Abstract: The paper presents some aspects of the (gray level) image binarization methods used in artificial vision systems. It is introduced a new approach of gray level image binarization for artificial vision systems dedicated to industrial automation – temporal thresholding. In the first part of the paper are extracted some limitations of using the global optimum thresholding in gray level image binarization. In the second part of this paper are presented some aspects of the dynamic optimum thresholding method for gray level image binarization. Starting from classic methods of global and dynamic optimal thresholding of the gray level images in the next section are introduced the concepts of temporal histogram and temporal thresholding. In the final section are presented some practical aspects of the temporal thresholding method in artificial vision applications form the moving scene in robotic automation class; pointing out the influence of the acquisition frequency on the methods results.

## 1. GLOBAL THRESHOLD FOR IMAGE BINARIZATION

Threshold methods are defined as starting from the analyze of the values of a function $T$ of the type:

$$T = T[x, y, p(x, y), f(x, y)] \quad (1)$$

Where:
$f(x, y)$ – represents the intensity value of the image element located on the coordinates $(x, y)$;
$p(x,y)$ – represents the *local properties* of the specific point (like the average intensity of a region centered in the coordinates $(x, y)$).
$T$ – is the *binarization threshold*
The goal is to obtain from an original gray level image, a binary image $g(x, y)$ defined by:

$$g(x, y) = \begin{cases} 1 & \text{for } f(x, y) > T \\ 0 & \text{for } f(x, y) \leq T \end{cases} \quad (2)$$

For $T$ a function only of $f(x, y)$, the obtained threshold is called *global threshold*.
In the case of $T$ a function of both $f(x, y)$ and $p(x, y)$, the obtained threshold is named *local threshold*.
In the case of $T$ a function of all $f(x, y)$, $p(x, y)$, $x$ and $y$, the threshold is a *dynamic threshold*.

### 1.1. Intensity level normal distribution assumption

*Gray level histogram* represents the probability density function of the intensity values of the image. In order to simplify the explanations, we suppose the image histogram of the gray levels is composed from two values combined with additive Gaussian noise:
- The first segment of the image histogram corresponds to the background points – the intensity levels are closer to the lower limit of the range (the background is dark)
- The second segment of the image histogram corresponds to the object points – the intensity levels are closer to the upper limit of the intensity range (the objects are bright).

The problem is to estimate a value of the threshold $T$ for which the image elements with an intensity value lower than $T$ will contain background points and the pixels with the intensity value greater than $T$ will contain object points, with a minimum error.

For a real image, the partitioning between the two brightness levels (the background and the object) is not so simple and also not so accurate.
The partitioning is fully accurate only if the two modes of the bimodal histogram are not overlapped.
In Fig.1 is presented a possible shape of a bimodal histogram with overlapped modes.

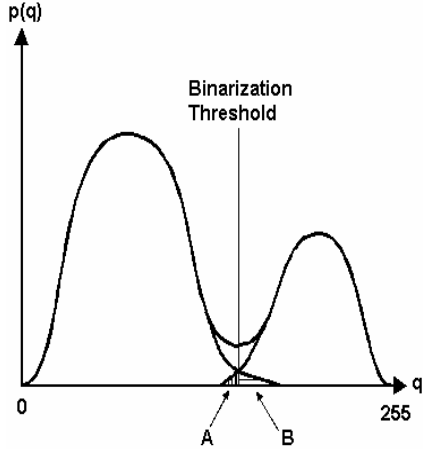

Fig.1 Bimodal histogram with overlapped modes

The classification is defined as the process of the distribution of the pixels in classes (the class of the object pixels and the class of background pixels).
The goal of the binarization process is the minimization of the error of classification. The optimum binarization threshold is located in the intersection position of the two normal distributions.
The estimation of the error of classification is obtained from the area of the overlapped segments:

$$E = \frac{A+B}{\text{image size}} \quad (3)$$

Suppose the image contains two intensity level values affected with additive Gaussian noise. The mixture probability density function is:

$$p(x) = P_1 p_1(x) + P_2 p_2(x) \quad (4)$$

Where:
$x$ – the random value representing the intensity level,
$p_1(x)$, $p_2(x)$ – are the probability density functions,
$P_1$, $P_2$ - are the a priori probabilities of the two intensity levels ($P_1 + P_2 = 1$).
For the normal distribution case on the two brightness levels:

$$p(x) = \frac{P_1}{\sqrt{2\pi}\sigma_1} \exp\frac{(x-\mu_1)^2}{2\sigma_1^2} + \frac{P_2}{\sqrt{2\pi}\sigma_2} \exp\frac{(x-\mu_2)^2}{2\sigma_2^2}$$
(5)

Where:

$\mu_1, \mu_2$ - are the mean values of the two brightness levels (the two modes),
$\sigma_1, \sigma_2$ - are the standard deviations of the two statistical populations.
Suppose the background is darker than the object. In this case $\mu_1 < \mu_2$ and defining a threshold $T$, so that all pixels with intensity level below $T$ are considered belonging to the background and all pixels with level above $T$ are considered object points. The probability of misclassification an object point (classifying an object point as a background point) is:
Similarly, $E_2$:

$$E_1(T) = \int_{-\infty}^{T} p_2(x)dx$$
$$E_2(T) = \int_{T}^{+\infty} p_1(x)dx \quad (6)$$

The probability of error is given by:

$$E(T) = P_1 E_2(T) + P_2 E_1(T) \quad (7)$$

To find the threshold value for which the error is minimum, $E(T)$ is differentiate with respect to $T$:

$$P_1 p_1(t) = P_2 p_2(t) \quad (8)$$

Applying the result to the Gaussian density we obtain:

$$AT^2 + BT + C = 0, \quad (9)$$

Where:

$$A = \sigma_1^2 - \sigma_2^2 \quad (10)$$
$$B = 2(\mu_1 \sigma_2^2 - \mu_2 \sigma_1^2)$$
$$C = \sigma_1^2 \mu_2^2 - \sigma_2^2 \mu_1^2 + \sigma_1^2 \sigma_2^2 \ln\frac{\sigma_1 P_1}{\sigma_2 P_2}$$

If the standard deviations are equal, a single threshold is sufficient:

$$T = \frac{\mu_1 + \mu_2}{2} + \frac{\sigma^2}{\mu_1 - \mu_2} \ln\frac{P_2}{P_1} \quad (11)$$

If the probabilities are equal $P_1 = P_2$, the threshold value is equal with the average of the means.
A way of checking the validity of the assumption of bimodal histogram is to estimate the mean-square error between the mixture density, $p(x)$ and the experimental histogram $h(x_i)$.

$$M = \frac{1}{N}\sum_{i=1}^{N}[p(x_i) - h(x_i)]^2 \quad (12)$$

Where: $N$ – number of possible levels of the image (usually $N = 256$)

The image binarization is obtained changing the color attribute of each pixel according to its intensity level relative to the binarization threshold.

Characteristics of the global thresholding methods:
- The assumption that both classes have the same standard deviation is acceptable, but the assumption the classes (two levels) have the same a priori probabilities in many applications is not acceptable.

In the case of the artificial vision systems dedicated to object recognition for industrial applications there is a large amount of a priori information about the image that has to be processed.

Can be obtained better results of estimation of the distribution (standard deviation and the mean value) of the image elements of the scene (background image – without the objects).

Usually, in robotic applications, the illumination environment is known and controlled and also the object classes with a probability of apparition in the image are not known.

In many robotic application an estimation of the ratio between the area of the objects to be analyzed and the total area of the image scene, can be made with good results (a batter estimation than the assumption of $P_1 = P_2 = 0.5$).

## 2. DYNAMIC THRESHOLD FOR IMAGE BINARIZATION

There are some classes of scenes of artificial vision systems where using the global threshold methods is not acceptable:
- The case of the applications where the lighting system does not supply a uniform intensity all over the analyzed surface.
- Segments of the image (or some times, image elements) do not have the same behavior in the same lighting conditions.

For these types of images, for binarization of the image, the most often used are dynamic threshold methods.

The algorithm of the estimation of the dynamic threshold consist of:
The methods are based on the local analyze of the image.
- The original image is divided in regions of a prescribed size.
- For each region it is estimated the histogram
- For each histogram it is estimated the error induced from the assumption of bimodal histogram (a histogram built from two normal distributions)
    - If the value of the error is less than an acceptable value, the global threshold for the region is estimated.
    - If the value of the error is too big (the histogram is too far from a bimodal histogram) the threshold value for binarization is estimated from the interpolation of the neighbors region threshold values (for which the assumption of a bimodal histogram is considered acceptable).
- In the final stage, a second interpolation process is applied: for each image element is assigned a threshold value $T(x, y)$ from the interpolation of the values of the neighbor image elements.

The method is called dynamic thresholding because the value of the resulted threshold for each image element is dependent of the position of the element in the image - $T(x, y)$.

Characteristics of the dynamic thresholding methods:
- Lack of processing time consumption – each element of the image is used at least two times (the method requires multiple-pass of the image) in different steps of the algorithm (and the number of the elements is very large).
- Estimation of the acceptable error value (or the validation of the bimodal histogram assumption) is a complex process.
- To choose the size of the image regions we have to take into account:
    - Large size of the region makes the method to loose the dynamic threshold characteristics and to fail into a global threshold method
    - Small size of the region makes to loose the statistical characteristic of the population of the image elements contained by the analyzed region (and the accuracy of the results is lost).
- The last comment on the method is the fact that this method does not solve the problem of the non-uniformity of the illumination system or of the acquisition sensor.

## 3. TEMPORAL HISTOGRAM

For the class of artificial vision systems dedicated for moving scene (used very often in inspection and robotic applications) three types of image intensity level distortions can be identified:
- Illumination nonuniformity (obtaining a uniform intensity of the light on the whole area of the scene where the image is analyzed – usually 2 m – it is practical impossible).
- Sensor non-linearity – for linear cameras with a large number of pixels per row (2048 and more) can be identified areas of nonlinear behavior of the sensor (there are segments of the linear sensor with a different behavior of the elements sensitivity at light intensity).
- Sensor cells nonuniformity – in cameras with CCD sensor, the cells presents a different response on sensitivity at light intensity related to their neighbors

In Fig.2 are presented the image intensity level distortions.

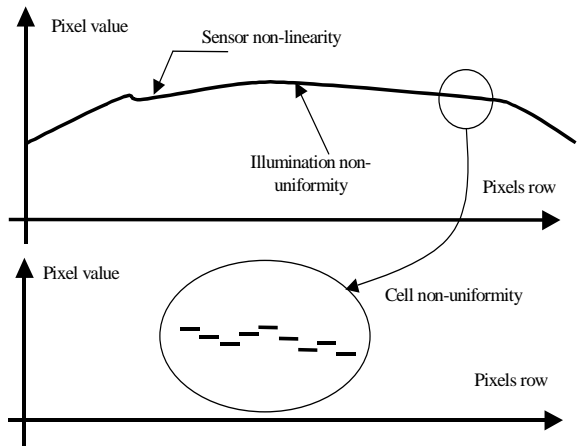

Fig.2 Image intensity level distortions for CCD linear camera acquisition

The main problem of the methods presented before represents the assumption that the image is a statistical population obtained from the addition of two ore more distributions (in the general accepted case normal distribution).

In the general case (an *array image*) an image represents a data set of:

$$\{f(x, y) \mid x \in [0,N], y \in [0,M]\}, \qquad (13)$$

where:
$N$ represents the number of image elements per row (number of image columns),
$M$ represents the number of image elements per column (number of image rows),

In the *linear image* case, this data set become:

$$\{f(x) \mid x \in [0,N]\}, \qquad (14)$$

where:
$N$ represents the number of image elements per row (number of image columns),

This assumption on the distribution of the intensity levels has the starting point the assumption that the insertion point of the noise is located on the transmission level of the information. In other words, the assumptions is that:
- The acquired image is an ideal image (with only two gray levels: the gray level of the scene pixels and the gray level of the pixels corresponding to the object)
- Then a global noise is applied, transforming the two levels in two normal distributions.
The assumption is false and using it we are analyzing a histogram, which is far away of two normal distributions, and from here the results are distorted.

In reality the noise on the intensity level has its insertion point on the acquisition level (noise on the "intensity source", sampling, holding and conversion, etc) and not on image transmission level. Intensity source has the meaning of intensity signal on the acquisition element and not only the lighting system. This implies the fact that the noise on the intensity source represents the whole chain of: lighting source noise, reflective characteristics of the object surface and reflective characteristics of the scene surface and the sensitivity characteristics of the sensor.

Moving the insertion point of the noise we obtain:
In the general case (an *array image*) an image represents a data set of:

$$\{f_i(x, y) \mid i \in [0,L]\}, \qquad (15)$$

where:
$L$ represents the number of the image frames (the size of the statistic population analyzed),
$x \in [0,N]$, $N$ representing the number of image elements per row (number of image columns),
$y \in [0,M]$, $M$ representing the number of image elements per column (number of image rows).

In the *linear image* case, this data set become:

$$\{f_i(x) \mid i \in [0,L]\}, \qquad (16)$$

where:
$L$ represents the number of the image frames (the size of the statistic population analyzed),
$x \in [0,N]$, $N$ representing the number of image elements per row (number of image columns).

In this way several temporal built statistical populations (from intensity levels of the same image element on a set of image frames acquired on different moments) replace the spatial built statistical population (made from image elements of the same image).

The method of temporal histogram has the result the fact that each element of this set of histograms represents a bimodal histogram with two not overlapped modes (in case of a correct acquisition environment).
It can be also introduce an estimation of the quality of the acquisition and binarization process using the estimation of the misclassification error analyzing the parameters of the two normal distributions (one corresponding to the object and the other one corresponding to the scene).

The method offers also the capacity of identification of the areas where some modifications should be done (on the lighting system) in order to improve the quality of the acquisition and binarization process.

The lack of the proposed method is the memory consumption (it has to be built $N \times M$ different histograms in array acquisition, or $N$ – in linear acquisition case). This problem is not so restrictive because at the end only the threshold values have to be stored and not the whole histograms.

Another restriction is the fact that the method requires a large number of image frames acquired for construction of the statistical populations (in application setup time). In the case of the systems dedicated to industrial applications usually this does not represent a real problem. This type of applications does not require a system response in condition of a small number of image frames a priori acquired (but other type of applications has these requirements).

The vision systems dedicated to industrial applications can take the advantage on the fact that the image environment does not change a lot in time. In this way it can be initially reserved a certain time (before the runtime of the system) for acquiring a large enough number of image frames in order to be able to identify the permanent characteristics of the environment. All the intensity level distortions (described in this section) presents permanent characteristics.

Using this method is a necessity for the artificial vision systems dedicated to applications where the errors on binarization are not acceptable (inspection and measurement systems – error free systems). In the applications dedicated exclusively to shape recognition the errors are accepted in a predefined range.

## 4. THE INFLUENCE OF THE ACQUISITION FREQUENCY ON THE BINARIZATION THRESHOLD VALUE

In moving scene applications, in order to maintain a constant resolution of the vision system along the direction of the scene movement, it is necessary the ratio between the acquisition frequency (the image lines rate – in the case of a line scan camera) and the scene speed to be constant.

The acquisition frequency determines the exposure time of the CCD sensor cells. It can be notice an important influence of the speed (of the conveyor) on the intensity level of the same image element in the same lighting environment.

In Fig.3 are presented the experimental results obtained analyzing the influence on the intensity levels (for both: bright object and dark background) of the speed of the conveyor (acquisition frequency). The results were obtained on a statistical population from an image element on each measured speed.
The second column represents the measured speed of the scene (conveyor) – V [m/min].
The $3^{rd}$ to $8^{th}$ columns represents image intensity levels (clarified in Fig.) estimated from the analyzed statistical population (temporal histogram).
The values from the Threshold column are the binarization threshold values obtained from a global optimum temporal thresholding method applied on the histogram built for each analyzed level of the speed.

In Fig.4 are presented graphical the explanations on the meanings of the data involved in the analysis of the influence of the speed (acquisition frequency) on the intensity levels.

The artificial vision system benefits from these results using a relation between the value of the binarization threshold and the speed V of the scene (acquisition frequency)

$$T = T(x, V) \qquad (17)$$

Because of the response time restrictions imposed to the artificial vision system, instead of using an explicit expression of the estimated function $T(x, V)$, a search method in an a priori filled table (at setup time) is more appropriate.

The size of the table is 256 (the number of the possible values of the binarization thresholds), containing floating-point values of the speed of the conveyor (acquisition frequency) for which the value of the binarization threshold has to be changed.

| V [m/min] | Object Min+ | Object Max | Object Min- | Scene Min+ | Scene Max | Scene Min- | Threshold |
|---|---|---|---|---|---|---|---|
| 1 | 20.7 | 221.532847 | 203.810219 | 168.364964 | 66.459854 | 44.306569 | 22.153285 | 110.766423 |
| 2 | 25.9 | 174.338624 | 160.391534 | 132.497354 | 52.301587 | 34.867725 | 17.433862 | 87.169312 |
| 3 | 31.1 | 147.510373 | 135.709544 | 112.107884 | 44.253112 | 29.502075 | 14.751037 | 73.755187 |
| 4 | 36.2 | 130.479452 | 120.041096 | 99.164384 | 39.143836 | 26.095890 | 13.047945 | 65.239726 |
| 5 | 41.3 | 118.513120 | 109.032070 | 90.069971 | 35.553936 | 23.702624 | 11.851312 | 59.256560 |
| 6 | 46.4 | 109.644670 | 100.873096 | 83.329949 | 32.893401 | 21.928934 | 10.964467 | 54.822335 |
| 7 | 51.4 | 102.927928 | 94.693694 | 78.225225 | 30.878378 | 20.585586 | 10.292793 | 51.463964 |
| 8 | 56.5 | 97.474747 | 89.676768 | 74.080808 | 29.242424 | 19.494949 | 9.747475 | 48.737374 |
| 9 | 61.6 | 93.040293 | 85.597070 | 70.710623 | 27.912088 | 18.608059 | 9.304029 | 46.520147 |
| 10 | 66.7 | 89.363484 | 82.214405 | 67.916248 | 26.809045 | 17.872697 | 8.936348 | 44.681742 |
| 11 | 72.5 | 85.877863 | 79.007634 | 65.267176 | 25.763359 | 17.175573 | 8.587786 | 42.938931 |

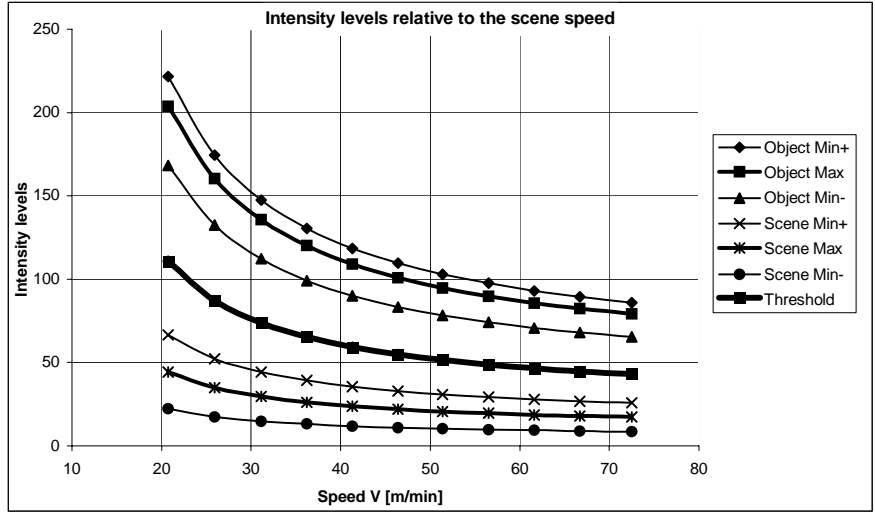

Fig.3 The influence on the intensity levels of the speed of the scene (acquisition frequency)

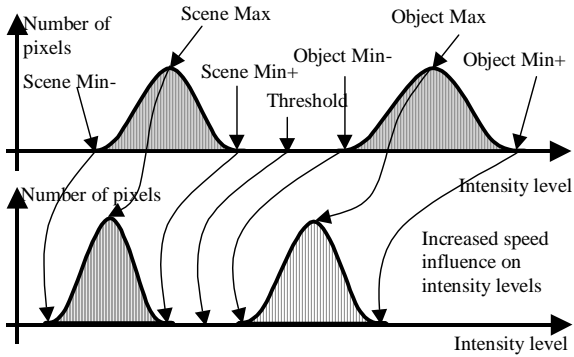

Fig.4 The influence of the speed on the intensity levels

## 5. CONCLUSIONS

For the class of artificial vision systems dedicated for inspection and measurement industrial applications the error on binarization process is not acceptable. In this, case classic methods like global, local and dynamic threshold are not applicable. The paper introduces a new approach of gray level image binarization – temporal thresholding.

For the class of artificial vision systems dedicated for moving scene the acquisition frequency is dependent on the speed of the transmission support (usually a conveyor). To solve this problem, the artificial vision system has to estimate the influence of the acquisition frequency on the histogram and on the binarization threshold values.

The paper proposes a processing time efficient method to estimate the binarization threshold for the case of an *error free* vision system in the case of variation of the acquisition frequency.